\pdfoutput=1

\documentclass[11pt]{article}

\usepackage{acl}
\usepackage{times}
\usepackage{latexsym}\usepackage{makecell}
\usepackage{bm}
\usepackage[T1]{fontenc}

\usepackage[utf8]{inputenc}
\usepackage{subfigure}
\usepackage{microtype}
\usepackage{enumitem}
\usepackage{graphicx}
\usepackage{multirow}
\usepackage{amsmath}
\usepackage{amssymb}
\usepackage{algorithm}
\usepackage{algorithmicx}
\usepackage{algpseudocode}
\usepackage{xcolor}
\usepackage{array}
\usepackage{booktabs}
\usepackage{arydshln}

\title{BatchEval: Towards Human-like Text Evaluation}
\author{Peiwen Yuan$^1$, Shaoxiong Feng$^2$, Yiwei Li$^1$, Xinglin Wang$^1$, Boyuan Pan$^2$\\ {\bf Heda Wang$^2$, Kan Li$^{1}$\footnotemark[1]}\\
  $^1$School of Computer Science and Technology, Beijing Institute of Technology \\
  $^2$Xiaohongshu Inc \\
  \texttt{\{peiwenyuan,liyiwei,wangxinglin,likan\}@bit.edu.cn} \\
  \texttt{\{shaoxiongfeng2023,whd.thu\}@gmail.com} \ \  \texttt{\{panby\}@zju.edu.cn}}
  
\begin{document}
\maketitle
\renewcommand{\thefootnote}{\fnsymbol{footnote}} 
\footnotetext[1]{Corresponding author.} 
\renewcommand{\thefootnote}{\arabic{footnote}}
\begin{abstract}
Significant progress has been made in automatic text evaluation with the introduction of large language models (LLMs) as evaluators. 
However, current sample-wise evaluation paradigm suffers from the following issues: 
(1) Sensitive to prompt design; 
(2) Poor resistance to noise; 
(3) Inferior ensemble performance with static reference. 
Inspired by the fact that humans treat both criterion definition and inter sample comparison as references for evaluation, we propose \textsc{BatchEval}, a paradigm that conducts batch-wise evaluation iteratively to alleviate the above problems. 
We explore variants under this paradigm and confirm the optimal settings are two stage procedure with heterogeneous batch composition strategy and decimal scoring format.
Comprehensive experiments across 3 LLMs on 4 text evaluation tasks demonstrate that \textsc{BatchEval} outperforms state-of-the-art methods by 10.5\% on Pearson correlations with only 64\% API cost on average.
Further analyses have been conducted to verify the robustness, generalization, and working mechanism of \textsc{BatchEval}\footnote{Our code and data have been released on \url{https://github.com/ypw0102/BatchEval}.}.

\end{abstract}
\section{Introduction}
Accurately evaluating the text quality of specific criterion (e.g., coherence) can facilitate better understanding, application, and development of large language models (LLMs), which becomes more crucial with their recent rapid progress in text generation capabilities \citep{GPT4}. 
Due to the labor-intensive and time-consuming nature of human evaluation, early works have explored automatic evaluation methods, which can be categorized into rule-based \citep{bleu,meteor}, embedding-based \citep{vextrema,bertscore}, and learning-based \citep{usr,mme_crs} approaches. 
Continuous progress has been achieved through these methods, but there remains a significant gap in their consistency with human judgments \citep{survey}.

\begin{figure}[t]
\centering
\includegraphics[width=0.48\textwidth]{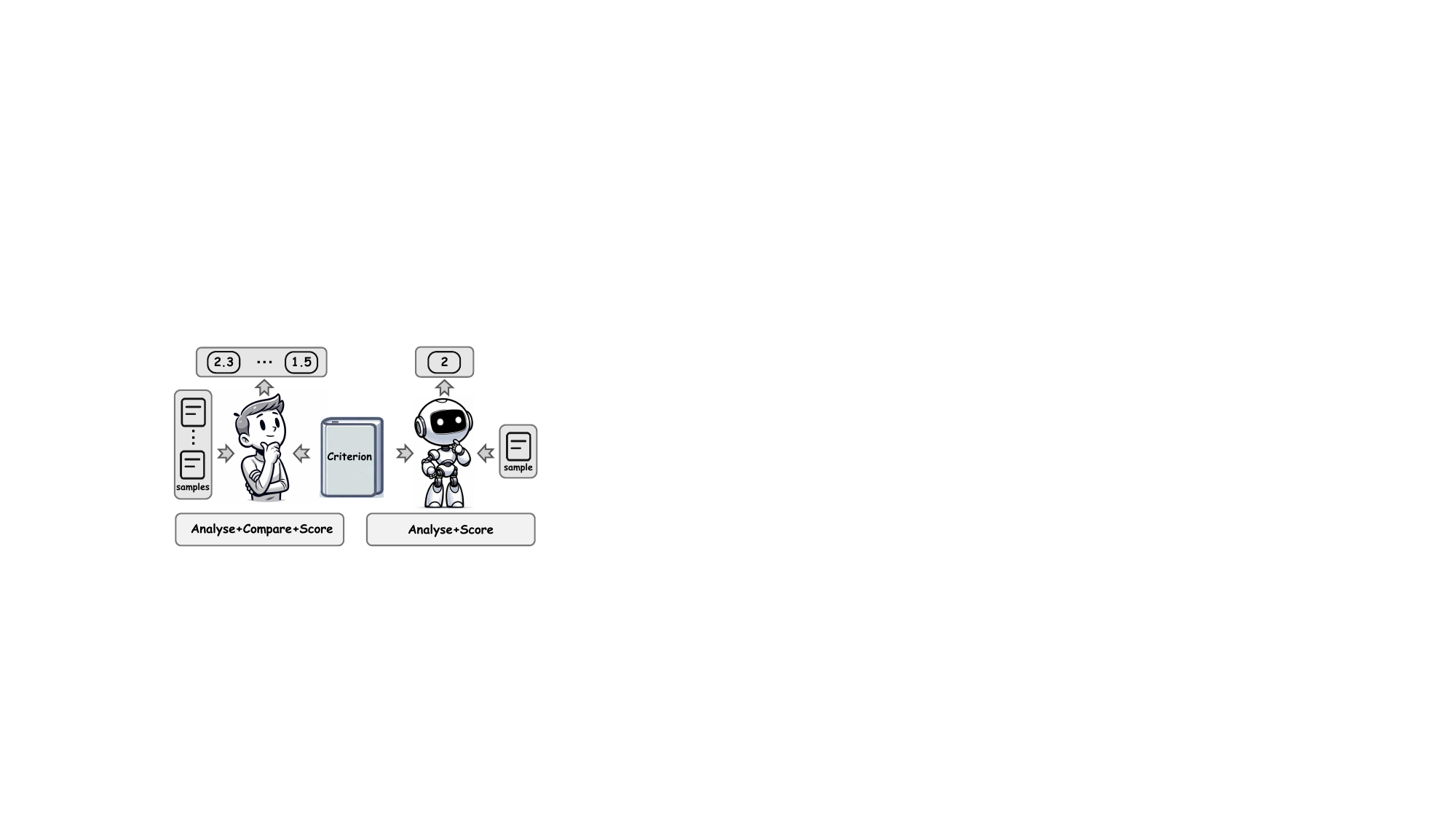} 
\caption{Human evaluators evaluate text quality based on criterion definition and sample comparison, while current LLM-based evaluators only rely on criterion.}
\label{fig:abs}
\end{figure}

Recently, the revolutionary power of LLMs has been applied across various fields, demonstrating performance that is even on par with humans \citep{GPT4,llm_survey}. 
In text evaluation filed, LLM-based evaluators \citep{canllm,geval,chateval,closer} have also made significant progress compared to traditional methods, but they still lag behind human evaluators. 
We carefully compared their working procedures and found that the difference in evaluation references might be the reason for the performance disparity (Figure~\ref{fig:abs}). 
Human evaluators analyze samples based on the criterion definition and provide discriminative scores through comparison between samples. 
However, LLM-based evaluators assess each sample individually, thus only having criterion as a reference.

We analyze that such a sample-wise evaluation paradigm will face problems on three aspects: 
(1) \textit{Robustness against prompt design.} 
Since criterion is the sole reference for evaluation, minor changes to the prompt may lead to significant variations in the evaluation results (See \S \ref{sec:rob_prompt} for empirical validation).
(2) \textit{Robustness against noise.}
Due to the absence of comparison between samples, the evaluation scores lack discrimination and exhibit a non-uniform distribution (See Figure~\ref{fig:dis}), which can lead to reduced robustness against noise\footnote{e.g., random deletion and synonym substitution on samples} (See Theorem~\ref{theorem1}).
(3) \textit{Performance under ensemble.}
Current LLM-based evaluators average scores from multiple generations as the final rating for given sample. However, generating multiple times from the static reference (criterion) induces a lack of diversity among scores (Figure~\ref{fig:diver}), which can weaken the effect of ensemble according to Theorem~\ref{theorem2}.

To address the aforementioned problems, we propose \textsc{BatchEval}, a new LLM-based text evaluation paradigm that assesses samples batch-wise, akin to the way of humans. 
Overall, \textsc{BatchEval} iterates through a process where all samples are split into several batches, with each batch then being compiled into a prompt for input to the LLM.
By introducing in-batch samples as an additional reference apart from criterion, the orthogonal and complementary references can not only reduce the dependency on prompt design but also enhance the discrimination of scores between samples through in-batch comparison, leading to improved robustness. 
Furthermore, the iteratively changing batch composition can provide LLMs with varying evaluation references, thereby enhancing diversity and the ensemble performance.

While the idea of \textsc{BatchEval} is simple, there are many ways it can be realized. 
We explored variants in evaluation procedure, format of scoring and composition of batch. 
Some of them work surprisingly well while some do not meet expectations.
Experiments and analyses confirm that separate analyzing and scoring evaluation procedure, decimal scoring format, and quality-heterogeneous batch composition strategy yield the optimal results.

We conduct extensive experiments on 4 text evaluation tasks primarily with GPT-4: turn-level response, dialogue, text summarization, and story generation. 
By allowing in-batch samples to share single prompt and applying a small iteration rounds, \textsc{BatchEval} outperforms best performing LLM-based evaluators by a significant margin (10.5\%) in terms of correlation with human evaluations, while incurring only 64\% of API costs.
We also validate the generalization of \textsc{BatchEval} on more LLMs, robustness to prompt design and noise, and analyze the choice of hyperparameters through further experiments. 
Finally, we probe into the working mechanism of \textsc{BatchEval} through attention analysis on Llama-2-70b-chat-hf. Our contributions are summarized as follows:

\begin{enumerate}

\item We analyzed how the sample-wise evaluation paradigm of LLM-based evaluators, differing from human evaluators, limited their robustness and consistency with human judgment.

\item We proposed \textsc{BatchEval}, a new paradigm that evaluates texts batch-wise, and experimentally validated its optimal settings.

\item We validated through experiments on 4 tasks that \textsc{BatchEval} outperforms public state-of-the-art methods by 10.5\% while incurring only 64\% of the API cost.

\item We analyzed the generalization, robustness, hyperparameter selection, and probed into the working mechanism of \textsc {BatchEval}.

\end{enumerate}

\section{Background}

\subsection{Automatic Text Evaluation}
Automatic text evaluation method has been extensively studied as a supplement to labor-intensive and time-consuming human evaluation, with its correlation to human judgment as the criterion for assessment. 
Both \textit{rule-based} \citep{bleu,meteor} and \textit{embedding-based} \citep{bertscore,vextrema} evaluation methods rely on the assumption that high-quality generated texts should have a significant word overlap with reference texts. However, this assumption conflicts with the high entropy nature of text generation, restricting its consistency with humans. 
\textit{Learning-based} methods consider directly assessing text quality through supervised \citep{adem,dae} and self-supervised \citep{usr,mme_crs} approaches and achieve significant progress.
Recently, \textit{LLM-based} evaluators \citep{chateval,closer,geval} have demonstrated advanced consistency with humans leveraging their incredible knowledge and capabilities.
However, typical sample-wise evaluation paradigm of the above methods leads to a lack of inter-sample comparison during scoring process, which serves as an important reference for human evaluators.
Therefore, we propose \textsc{BatchEval} to fill this gap for better alignment with humans.

\subsection{The Theorems Involved}
\newtheorem{theorem}{Theorem}

\begin{theorem}
The robustness against noise correlates positively with the uniformity of evaluator scoring distribution. (See Appendix~\ref{sec:proofoft1} for derivation in details)
\label{theorem1}
\end{theorem}
\citet{bcr} proposed this theorem and verified that learning-based evaluators, by adjusting the training loss function to uniformize the score distribution, can achieve better robustness against noise. We have experimentally proven that sample-wise LLM-based evaluators also exhibit an uneven score distribution (Figure~\ref{fig:dis}), which can weaken their robustness against noise (Appendix~\ref{sec:robustness_noise}). Thus, we propose \textsc{BatchEval} for a more uniform score distribution and better robustness against noise.

\begin{theorem}
Given scores from multiple generations of certain LLM $\mathcal{S}=\{s_i|i=1,..,N\}$ and human evaluation score $y$ for sample $x$, $\bar{s}$ is the average of $\mathcal{S}$, the following equation holds:
\begin{equation}
\small
Err(\bar{s},y) = Err(\mathcal{S},y)-Var(\mathcal{S}) 
\label{equation: div-re}
\end{equation}
where:
\begin{equation}
\setlength{\jot}{2pt}
\small
\begin{gathered}
Err(\bar{s},y) = (\bar{s}-y)^2\\
Err(\mathcal{S},y) = \frac{1}{N} \sum_{i=1}^{N} (s_i-y)^2\\
Var(\mathcal{S}) = \frac{1}{N} \sum_{i=1}^{N} (s_i-\bar{s})^2
\end{gathered}
\label{equation: div-re-detail}
\end{equation}
\label{theorem2}
\end{theorem}
Eq.~\eqref{equation: div-re} \citep{diverres} implies that smaller average error in single prediction scores ($Err(\mathcal{S},y)$) and larger variance among multiple prediction scores ($Var(\mathcal{S})$) induce smaller error in ensemble score ($Err(\bar{s},y)$). However, current sample-wise LLM-based evaluators score multiple times based solely on static reference (criterion), resulting in smaller $Var(\mathcal{S})$ (Figure~\ref{fig:bias}). To address this, we propose iterative quality-heterogenized batch composition strategy for LLMs to score with unbiased varying references, thus increasing $Var(\mathcal{S})$ for lower $Err(\bar{s},y)$.

\section{Methodology}

\begin{figure*}[t]
\centering
\includegraphics[width=0.96\textwidth]{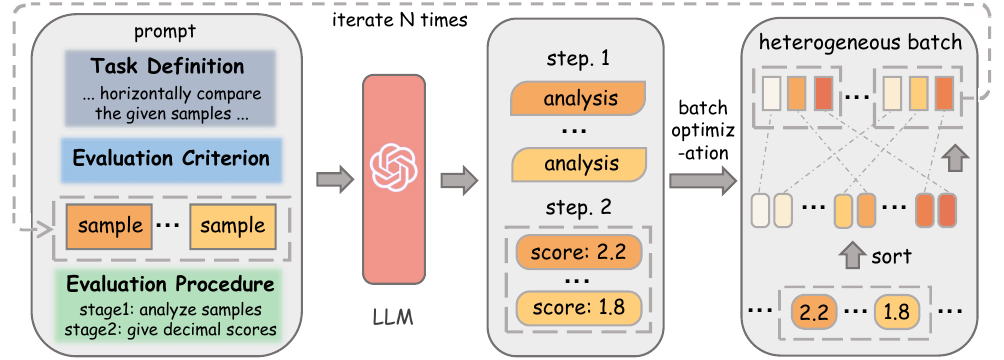} 
\caption{Overall illustration of \textsc{BatchEval}.}
\label{fig:main}
\end{figure*}

The core idea behind \textsc{BatchEval} is to fully use in-batch sample comparison to enhance evaluation accuracy and robustness. Algorithm \ref{alg:Bat} illustrates the working process of \textsc{BatchEval}, which involves $N$ rounds of iteration: (1) $B$ samples of each batch are compiled with pre-defined (task, criterion, evaluation procedure) into a single prompt for input to the LLM; (2) Based on the LLM's assessment of the samples' quality, we optimize batch allocation according to certain batch composition strategy. The core designs throughout the process are \textit{\textbf{how to evaluate}} (evaluation procedure), \textit{\textbf{what to input}} (batch composition strategy), and \textit{\textbf{what to output}} (scoring format). Below we discuss their potential variants in detail.

\begin{algorithm}
\small
    \caption{Workflow of \textsc{BatchEval}.}\label{alg:Bat}
        \begin{algorithmic}[1]
\Require Samples $x^{1:|\mathcal{D}|}$, LLM $\mathcal{M}$, Evaluation procedure $P$  \newline
        Task and criterion $T$, Iteration rounds $N$ , Batchsize $B$ \newline
        Batch composition strategy B{\fontsize{8}{12}\selectfont ATCH}S{\fontsize{8}{12}\selectfont TRATEGY}
\Ensure Ensemble evaluation scores $\Bar{s}^{1:|\mathcal{D}|}$
\State Randomly divide $x^{1:|\mathcal{D}|}$ into batches $b^{1:L}$, $L=\lceil \frac{|\mathcal{D}|}{B} \rceil$
\State $S_{all} \leftarrow \{i:[\ ]$ for $i$ $\in [1,|\mathcal{D}|]\}$
\For{$i\gets 1, N$}
    \State $S_{current} \leftarrow \varnothing$
    \For{$j\gets 1, L$}:
        \State $S_{current} \leftarrow S_{current}$.Append$(\mathcal{M}(T,P,b^j))$
    \EndFor
    \State $S_{all} \leftarrow S_{all}$.Merge$(S_{current})$
    \State $b^{1:L} \leftarrow $B{\fontsize{8}{12}\selectfont ATCH}S{\fontsize{8}{12}\selectfont TRATEGY}$(x^{1:|\mathcal{D}|},S_{all},B)$
\EndFor
\State $\Bar{s}^{1:|\mathcal{D}|} \leftarrow$Average$(S_{all})$
\end{algorithmic}
\end{algorithm}

\subsection{\textit{How to Evaluate}}
LLM-evaluators conduct sample-wise evaluation through a process of either analyzing followed by scoring \citep{chateval} or scoring followed by analyzing \citep{geval}, where the former typically performs better \citep{closer} possibly due to the effect of chain-of-thought \citep{COT}. Following this insight, we explored three possible evaluation procedures for \textsc{BatchEval} (See Appendix~\ref{sec:prompts} for prompts):
\paragraph{One stage} 
As the most intuitive extension of sample-wise evaluation, LLM analyzes and scores each sample of the batch in order. This procedure enables adequate comparison between samples, but insufficient comparison between analyses (the analyses of subsequent samples cannot be referenced by the earlier samples for scoring).

\paragraph{Two stage} 
To enhance the comparison among analyses, the LLM first analyzes all the samples. Based on the full comparisons among samples and analyses, the LLM further scores for each sample.

\paragraph{Three stage} 
From human experience, it can be easier to first rank and then score the samples, as compared to directly scoring them.
Therefore, we consider a procedure that sequentially performs analyzing, ranking, and scoring for all samples.

\subsection{\textit{What to Input}}
The composition of the batch largely determines the efficacy of in-batch comparison as evaluation reference.
One basic method is to fix a random batch division, attain multiple scores from the LLM and then average them for each sample. However, based on preliminary experiments and Theorem~\ref{theorem2}, we have found that this method does not yield good results due to the lack of diversity. Therefore, we consider redrawing the batch divisions after each round of evaluation to provide the LLM with varying references when assessing a certain sample.\footnote{See Appendix~\ref{sec:batchstrategydetail} for strategies in detail.}

\paragraph{Random Batch} 
One naive way is to reallocate batches randomly after each round of evaluation.

\paragraph{Homogeneous Batch}
Based on the idea of coarse-to-fine evaluation, we consider forming homogeneous batches in which samples have similar scores from the previous round of evaluation, in the hope that these samples can be further compared by LLM and ultimately attain discriminative scores.

\paragraph{Heterogeneous Batch} 
A contrary idea is to select samples with diversified scores based on the previous round of evaluation results to form a new batch. 
In this way, LLM develops an unbiased perception of samples with different qualities through batch optimization, thus scoring more accurately.

\subsection{\textit{What to Output}}
\label{sec:whatoutput}
Sample-wise evaluation methods typically apply integers as the format for LLM scoring \citep{geval,closer}, and \citet{llmeval} proved that using more refined scoring format can not bring additional gains. 
\textit{Will this trend be similar in \textsc{BatchEval}?}
Let us consider a concrete example: there are two samples with close but different quality, with human ratings of 2.2 and 1.8, respectively. 
Due to having only the criterion as reference, sample-wise evaluators may consider them to be close to the 2-point standard and consequently assign a score of 2 regardless of whether decimal is allowed. 
However, if they appear in the same batch, on the basis of judging that they are all close to 2 points, LLM can further compare their quality directly. 
Thus, it is possible for LLM to give them differentiated decimal scores if it is allowed, thereby achieving more consistent judgments with humans.
Based on the analysis above, we consider trying out two different scoring formats: \textbf{integer} and \textbf{decimal}.

Our default settings of \textsc{BatchEval} include two stage evaluation procedure, heterogeneous batch composition strategy and decimal scoring format, as shown in Figure~\ref{fig:main}.

\section{Experiments}
Centered around \textsc{BatchEval}, we will empirically explore the optimal variants in \S\ref{variants}, demonstrate its performance on different LLMs and tasks in \S\ref{overall}, validate the robustness in \S\ref{robustness}, and delve into its working mechanism in \S\ref{analysis}. We also investigate the choice of hyperparameters in Appendix \S\ref{sec:hyper}.

\begin{table*}[ht]
    \renewcommand\arraystretch{1.3}
    \small
    \centering
    \setlength{\tabcolsep}{0.08em} 
    \begin{tabular}{clcrcrcrcrcrcrrc}
    \toprule
    \multirow{2}{*}{\textbf{Type}} &\multirow{2}{*}{\textbf{Method}} & \multirow{2}{*}{\textbf{Scheme}\ } & \multicolumn{2}{c}{\textbf{\ \ Engaging\ \ }} & \multicolumn{2}{c}{\textbf{Understand}} & \multicolumn{2}{c}{\ \textbf{Naturalness}\ } & \multicolumn{2}{c}{\ \ \textbf{Coherence}\ \ }& \multicolumn{2}{c}{\ \ \ \textbf{Overall}\ \ \ \ }& \multicolumn{3}{c}{\ \ \ \ \ \ \ \ \ \ \textbf{Average}\ \ \ \ \ \ \ \ \ \ } \\
     && & $\bm{r_p}$\ &\ $\bm{r_s}$&$\bm{r_p}$\ &\ $\bm{r_s}$&$\bm{r_p}$\ &\ $\bm{r_s}$&$\bm{r_p}$\ &\ $\bm{r_s}$&$\bm{r_p}$\ &\ $\bm{r_s}$&$\bm{r_p}$\ &$\bm{r_s}$\ \ & \$$\bm{/item}$ \\   
    \hline
    Human&\multicolumn{2}{l}{Inter-annotator$^*$  }&.575&.581&.510&.510&.486&.487&.558&.560&.710&.718&.568&.571&-\\
    \hline
    \multirow{2}{*}{Rule} &\multirow{1}{*}{BLEU-4$^*$} & - & .232&.316&.201&.218&.180&.175&.131&.235&.216&.296&.192&.248&- \\
     &\multirow{1}{*}{METEOR$^*$} &  - & .367&.439&.245&.225&.212&.191&.250&.302&.337&.391&.282&.310&- \\
    \cline{1-16}
    \multirow{2}{*}{Embedding}&\multirow{1}{*}{V-Extrema$^*$} & - & .210&.205&.156&.132&.101&.076&.184&.184&.203&.209&.171&.161&- \\
    &\multirow{1}{*}{BERTScore$^*$} & - & .317&.335&.256&.226&.226&.209&.214&.233&.298&.325&.262&.266&- \\
    \cline{1-16}
    \multirow{2}{*}{Learning}&\multirow{1}{*}{USR$^*$} & - & .456&.465&.293&.315&.276&.304&.416&.377&.422&.419&.373&.376&- \\
    &\multirow{1}{*}{BCR} & - & .460&.463&.297&.325&.260&.298&.425&.391&.437&.421&.376&.380&- \\
     \cline{1-16}
    \multirow{8}{*}{LLM}&\multirow{1}{*}{G-Eval} & - & .710&.719&.568&.593&.595&.605&.576&.584&.717&.705&.633&.641&\textcolor{gray}{.0614} \\
    &\multirow{1}{*}{CloserLook} & - & .651&.688&.649&.699&.656&.665&.675&.687&.778&.772&.682&.702&\textcolor{gray}{.0686} \\
    &\multirow{1}{*}{CloserLook} & + ICL & .714&.743&.603&.685&.679&.693&.720&.733&.786&.783&.700&.727&\textcolor{gray}{.0856} \\
    \cline{2-16}
    &\multirow{6}{*}{\makecell[c]{\textsc{BatchEval}\\(Ours)}} &  one stage &\ .780&.783&\ .642&.680&\ .706&.710&\ .727&.729&\ .785&.793&.728&\ \ .739&\textcolor{gray}{.0525}\\
   & & three stage & .782&.778&.667&.725&.712&704&.712&.714&.797&.798&.734&.744&\textcolor{gray}{.0541}\\
    \cdashline{3-16}
   &  & random &.746&.743&.685&.724&.711&.700&.716&.720&.798&.799&.731&.737&\textcolor{gray}{.0528}
\\
  &  & homogeneous & .654&.663&.639&.607&.671&674&.669&.631&.722&.703&.671&.656&\textcolor{gray}{.0537}\\
    \cdashline{3-16}
  &  & integer & .771&.778&.686&\textbf{.732}&.726&.727&.722&.727&.790&.783&.739&.749&\textcolor{gray}{.0526}
 \\
  \cdashline{3-16}
  & &default & \textbf{.792}&\textbf{.790} & \textbf{.694}&.727 & \textbf{.730}&.\textbf{735} & \textbf{.740}&\textbf{.744}  & \textbf{.805}&\textbf{.800} & \textbf{.752}&\textbf{.759} & \textcolor{gray}{.0529} \\
    \bottomrule
    \end{tabular}
    \caption{Turn-level Pearson ($\bm{r_p}$) / Spearman ($\bm{r_s}$) correlations and average API cost per sample (\$$\bm{/item}$) of different metrics on Topical-Chat benchmark. The results of methods with $^*$ come from USR. We reproduced other methods with a unified API (the results were generally better than those reported in the original paper). All results of our replication are statistically significant (p-value < 0.05).}
    \label{tab: Main results}
\end{table*}

\subsection{Experimental settings}
\paragraph{Benchmarks}
A brief introduction of benchmarks involved are listed as follows:
\begin{itemize}[leftmargin=20pt]
\setlength{\itemsep}{0pt}
\setlength{\parsep}{0pt}
\setlength{\parskip}{0pt}
\item \textbf{Topical-Chat} \citep{usr} is a benchmark for evaluating dialogue response generation. To save costs, we exclude knowledge as input to LLM and therefore choose criteria where knowledge is not necessary: \texttt{Naturalness}, \texttt{Coherence}, \texttt{Engaging}, \texttt{Naturalness} and \texttt{Overall}.
\item \textbf{FED} \citep{fed} includes human ratings on 11 criteria to evaluate the quality of dialogue. We choose the top 4 important criteria as claimed in the original paper for evaluation: \texttt{Coherent}, \texttt{Understanding}, \texttt{Likeable} and \texttt{Overall}.
\item \textbf{HANNA} \citep{hanna} serves as a benchmark for meta-evaluating evaluation methods on story generation, with criteria including: \texttt{Coherence}, \texttt{Relevance}, \texttt{Empathy}, \texttt{Surprise}, \texttt{Engagement} and \texttt{Complexity}.
\item \textbf{QAGS} \citep{hanna} is a benchmark for evaluating the \texttt{Factual Consistency} of summaries on CNN \citep{cnn} and XSUM \citep{xsum}.
\end{itemize}

\begin{figure*}[t]
\centering
\includegraphics[width=0.96\textwidth]{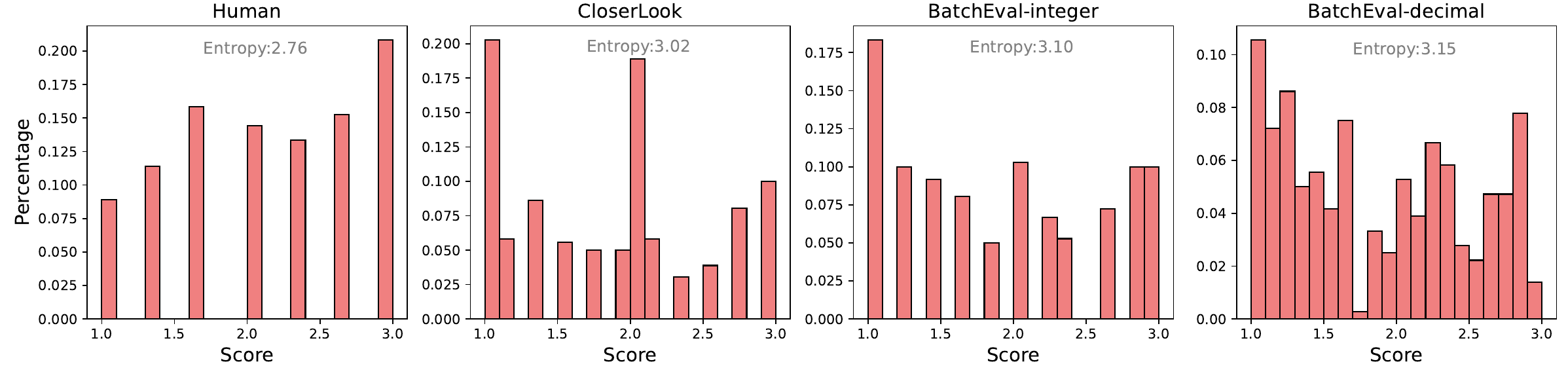} 
\caption{Score distribution and corresponding entropy ($-\sum_{s} p(s) \log_2 p(s)$) of different methods.}
\label{fig:dis}
\end{figure*}

\begin{figure}[t]
\centering
\includegraphics[width=0.48\textwidth]{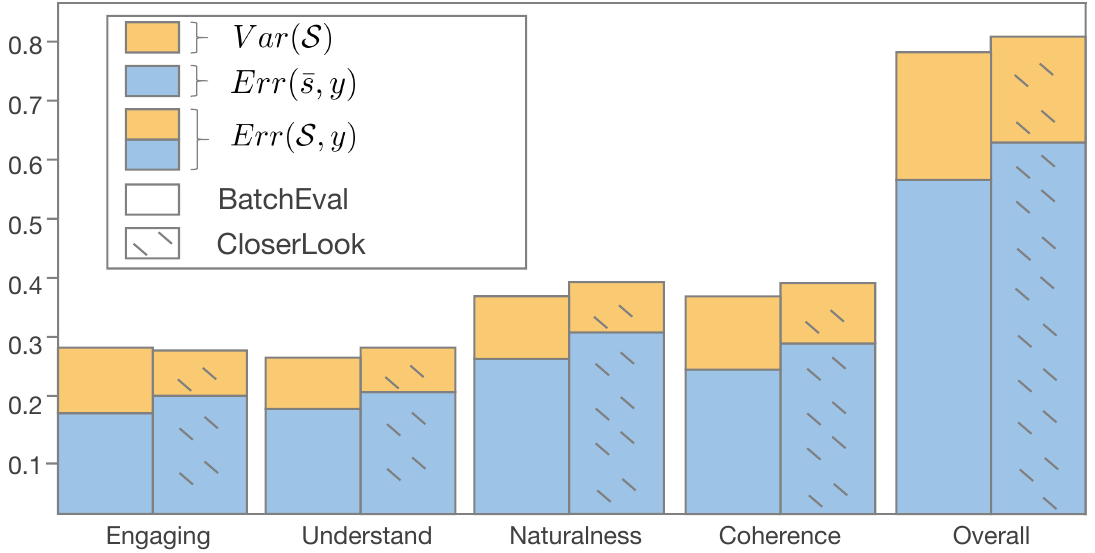} 
\caption{Comparisons between \textsc{BatchEval} and CloserLook from the perspective of Theorem~\ref{theorem2}.}
\label{fig:diver}
\end{figure}

\begin{figure}[t]
\centering
\includegraphics[width=0.48\textwidth]{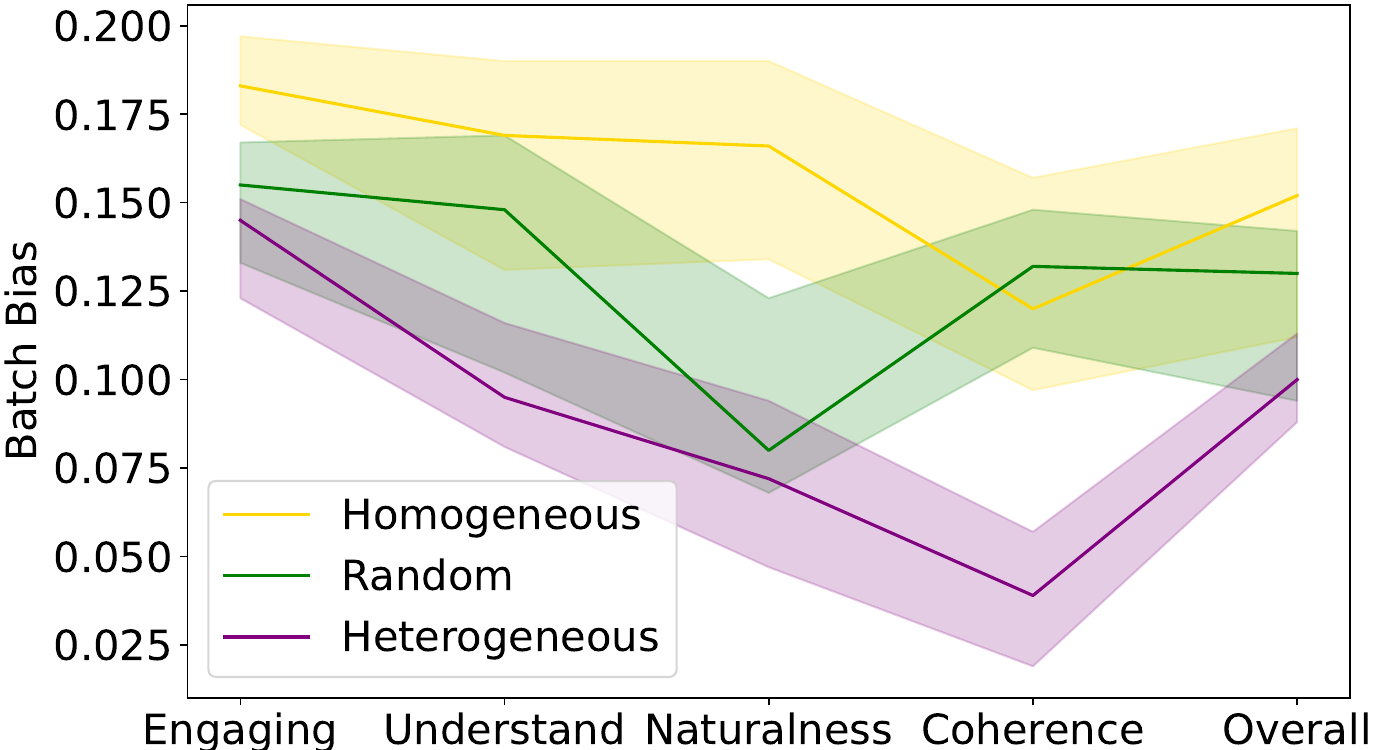} 
\caption{Average batch bias of different strategies.}
\label{fig:bias}
\end{figure}

\paragraph{Baselines}
We introduce four types of baseline methods in the experiments. 
Among them, both rule-based and embedding-based methods need reference text, which is unavailable in FED and QAGS. 
Learning-based methods are typically task-specific. Below we briefly list their categories and snapshots of LLM-based methods. 
Refer to Appendix~\ref{sec:baselines} for detailed introductions.
\begin{itemize}[leftmargin=20pt]
\setlength{\itemsep}{0pt}
\setlength{\parsep}{0pt}
\setlength{\parskip}{0pt}
\item \textbf{Rule-based}: BLEU \citep{bleu}, METEOR \citep{meteor}.
\item \textbf{Embedding-based}: Vector Extrema \citep{vextrema}, BERTScore \citep{bertscore}
\item \textbf{Learning-based}: USR \citep{usr}, BCR \citep{bcr}, FED \citep{fed}, DynaEval \citep{dynaeval}, QAGS \citep{qags}.
\item \textbf{LLM-based}\footnote{Two latest and well-known LLM evaluators are included. We are unable to reproduce some other methods due to incomplete disclosure of codes or prompts.}: G-Eval \citep{geval} recommended using LLM to evaluate according to the procedures generated by itself. \citet{closer} tried various evaluation schemes and proved through experiments that \textit{analyze-rate} led to the best performance, which we denote as CloserLook.
\end{itemize}

\paragraph{Details}
We explore variants of \textsc{BatchEval} on Topical-Chat for its wide recognition. 
If not specified, FED serves as our default dataset for exploratory experiments as it only has 125 samples, thus can save API expenses.
The other two benchmarks are used to confirm the generalization across tasks of \textsc{BatchEval}.
We primarily conduct experiments with GPT-4 (\textit{0613}) and validate the generalization across models of \textsc{BatchEval} with GPT-3.5-turbo (\textit{0613}) and Llama-2-70b-chat-hf. 
We set iteration rounds as 5, batchsize as 10, decoding temperature as 0.2 for all the experiment. 
For other LLM-based evaluators, we reproduced them according to their default settings (20 generations per sample) with the same API for a fair comparison. 
We choose Pearson and Spearman correlations to measure consistency with humans and also report API expenses for adequate comparison. We follow \citep{closer} to design prompts (See prompts in Appendix~\ref{sec:prompts}).

\subsection{Variants Exploration}
\label{variants}
As shown in Table~\ref{tab: Main results}, based on the default settings shown in Figure~\ref{fig:main}, we validate the effects of different variants (replacing the default setting with specific scheme) of \textsc{BatchEval}. 

\paragraph{Evaluation Procedure}
Compared to one stage procedure, the two stage procedure (default) achieves higher correlations by enhancing the comparison among analyses during scoring. Surprisingly, however, the three stage procedure does not perform well as expected. We speculate this may be due to the LLM's over-reliance on ranking results while neglecting the analyses and samples during scoring, and valide this in Appendix~\ref{sec:rank}.

\paragraph{Batch Composition Strategy}
As shown in Table~\ref{tab: Main results}, the performance of batch composition strategies ranks as follows: heterogeneous (default) > random > homogeneous. To investigate the reasons, we introduce batch bias as follows:
\begin{equation}
\small
Bias(\mathcal{B})= abs(\sum\limits_{i\in \mathcal{B}}s_i^\mathcal{B}-\sum\limits_{i\in \mathcal{B}}\bar{s}_i)/|\mathcal{B}|
\label{equation: bias}
\end{equation}
where $\mathcal{B}$ denotes the set of sample indexes of certain batch, $s_i^{\mathcal{B}}$ denotes score of sample $x_i$ generated with batch $\mathcal{B}$, $\bar{s}_i$ denotes average score of sample $x_i$ across all the iterations. Ideally, we aspire for the batch bias to approach zero. This implies that LLM should not have the overall scores in a batch skewed either high or low compared to the ensemble scores. We evaluate the average $Bias(\mathcal{B})$ of different strategies and find that $Bias(\mathcal{B})$ correlates negatively with correlations $\bm{r_s}$ and $\bm{r_p}$ (Figure~\ref{fig:bias}). This indicates that the more varied the quality of samples in a batch, the better they can simulate a real distribution as an unbiased reference to bring smaller batch bias for better correlations.

\paragraph{Scoring Format}
We observe from Table~\ref{tab: Main results} that decimal scoring format brings around 1 point correlations improvement upon integer. As shown in Figure~\ref{fig:dis}, the decimal scheme brings a more uniform scoring distribution. This implies that LLM indeed assigns more discriminative scores to different samples through in-batch comparison if decimal score is allowed, which verifies our hypothesis in \S\ref{sec:whatoutput} and accounts for the progress.

\begin{table*}[t]
    \renewcommand\arraystretch{1.3}
    \small
    \centering
    \setlength{\tabcolsep}{0.11em} 
    \begin{tabular}{clcrcrcrcrcrrc}
    \toprule
    \multirow{2}{*}{\textbf{Type}} &\multirow{2}{*}{\textbf{Method}} & \multirow{2}{*}{\textbf{Model}}& \multicolumn{2}{c}{\textbf{\ \ Likeable\ \ }} & \multicolumn{2}{c}{\textbf{Understand}} & \multicolumn{2}{c}{\ \textbf{Coherent}\ } & \multicolumn{2}{c}{\ \ \ \ \textbf{Overall}\ \ \ \ }& \multicolumn{3}{c}{\ \ \ \ \ \ \ \ \ \ \textbf{Average}\ \ \ \ \ \ \ \ \ \ } \\
     & & & $\bm{r_p}$\ \ &\ $\bm{r_s}$&$\bm{r_p}$\ \ &\ $\bm{r_s}$&$\bm{r_p}$\ \ &\ $\bm{r_s}$&$\bm{r_p}$\ \ &\ $\bm{r_s}$&$\bm{r_p}$\ \ &$\bm{r_s}$\ \ & \$$\bm{/item}$ \\   
    \hline
    Human&Inter-annotator&-&-&.838&-&.809&-&.809&-&.830&-& .822&-\\
    \hline
    \multirow{3}{*}{Learning}&\multirow{1}{*}{USR} &  - &  .245&.226&.182&.178&.170&.185&.284&.302&.220&.223&- \\
    &\multirow{1}{*}{FED} &    - &.248&.262&.295&.306&.262&.253&.460&.449&.316&.318&- \\
    &\multirow{1}{*}{DynalEval} &  - &  .389&.393&.379&.368&.399&.409&.484&.490&.413&.415&- \\
     \cline{1-14}
    \multirow{8}{*}{LLM}&\multirow{1}{*}{CloserLook} & Llama-2-70b&.525&.550&.574&\textbf{.611}&\textbf{.640}&.563&.634&.639&.593&.591& - \\
    &\multirow{1}{*}{\textsc{BatchEval}} & Llama-2-70b&\textbf{.537}&\textbf{.563}&\textbf{.619}&.597&.627&\textbf{.648}&\textbf{.722}&\textbf{.732}&\textbf{.626}&\textbf{.635}& - \\
    
    \cdashline{2-14}
    &\multirow{1}{*}{CloserLook} & GPT-3.5-turbo&.681&.666&.691&.605&.726&.724&.687&\textbf{.709}&.696&.676&\textcolor{gray}{.0022} \\
    &\multirow{1}{*}{\textsc{BatchEval}} & GPT-3.5-turbo&\textbf{.682}&\textbf{.674}&\textbf{.704}&\textbf{.708}&\textbf{.733}&\textbf{.730}&\textbf{.705}&.699&\textbf{.706}&\textbf{.703}&\textcolor{gray}{.0011} \\
    \cdashline{2-14}
    &\multirow{1}{*}{G-Eval } & GPT-4&.638&.692&.670&.625&.707&.721&.689&.652&.676&.673&\textcolor{gray}{.0667} \\
    &\multirow{1}{*}{CloserLook \textit{w} human prompt} & GPT-4&.658&.680&.701&.614&.739&.751&.715&.684&.703&.682&\textcolor{gray}{.0785} \\
    &\multirow{1}{*}{CloserLook \textit{w} GPT-4 prompt} & GPT-4&.632&.660&.678&.639&.725&.749&.723&.678&.690&.682&\textcolor{gray}{.0827} \\
    &\multirow{1}{*}{\textsc{BatchEval} \textit{w} human prompt} & GPT-4&.731&\textbf{.741}&.778&.696&.753&\textbf{.753}&.738&\textbf{.729}&.750&\textbf{.730}&\textcolor{gray}{.0314} \\
    &\multirow{1}{*}{\textsc{BatchEval} \textit{w} GPT-4 prompt} & GPT-4&\textbf{.736}&\textbf{.741}&\textbf{.780}&\textbf{.700}&\textbf{.784}&.749&\textbf{.748}&.727&\textbf{.762}&.729&\textcolor{gray}{.0314} \\
    \bottomrule
    \end{tabular}
    \caption{Dialog-level Pearson ($\bm{r_p}$) / Spearman ($\bm{r_s}$) correlations and average API cost per sample (\$$\bm{/item}$) on FED-dialog benchmark. We implemented and tested all the methods with p-value < 0.05.}
    \label{tab: Main results2}
\end{table*}

\begin{table*}[t]
    \renewcommand\arraystretch{1.3}
    \small
    \centering
    \setlength{\tabcolsep}{0.12em} 
    \begin{tabular}{lccccccccccccccc}
    \toprule
    \multirow{2}{*}{\textbf{Method}} &\multicolumn{2}{c}{\ \textbf{Coherence}\ \ \ \ } & \multicolumn{2}{c}{\ \textbf{Relevance}\ \ \ \ } & \multicolumn{2}{c}{\ \ \textbf{Empathy}\ \ \ \ \ \ }& \multicolumn{2}{c}{\ \ \textbf{Surprise}\ \ \ \ \ \ }& \multicolumn{2}{c}{\ \textbf{Engagement}\ \ \ \ }& \multicolumn{2}{c}{\ \textbf{Complexity}\ \ \ \ }& \multicolumn{3}{c}{\ \ \ \ \ \textbf{Average}\ \ \ \ \ \ \ \ \ } \\
     & \ \ $\bm{r_p}$&\ $\bm{r_s}$&\ \ $\bm{r_p}$&\ $\bm{r_s}$&\ \ $\bm{r_p}$&\ $\bm{r_s}$&\ \ $\bm{r_p}$&\ $\bm{r_s}$&\ \ $\bm{r_p}$&$\bm{r_s}$\ \ &\ \ $\bm{r_p}$&$\bm{r_s}$\ \ &\ \ $\bm{r_p}$\ \ \ &$\ \ \bm{r_s}$\ \ \ & \ \ \$$\bm{/item}$ \\   
    \hline

    \multirow{1}{*}{BLEU-4}  &.220&.218&.135&.175&.242&.216&.178&.224&.242&.270 &.362&.273&.230&.229&-\\
    \multirow{1}{*}{METEOR}  &.335&.273&.202&.190&.304&.282&.285&.283&.316&.338 &.520&.482&.307&.307&-\\
    \multirow{1}{*}{BERTScore} &   .358&.293&.201&.188&.308&.303&.302&.290&.308&.331&.501&.472&.330&.313&- \\
    \multirow{1}{*}{G-Eval} &    .572&.578&.582&.584&.453&.461&.311&.347&.562&591&.602&.557&.514&.520&\textcolor{gray}{.0772} \\
    
    \multirow{1}{*}{CloserLook} &    .595&.591&.579&.597&.498&.478&.280&.339&.605&\textbf{.607}&.619&.568&.529&.530&\textcolor{gray}{.0835} \\
    \multirow{1}{*}{\textsc{BatchEval}} &    \textbf{.678}&\textbf{.625}&\textbf{.702}&\textbf{.679}&\textbf{.546}&\textbf{.543}&\textbf{.368}&\textbf{.381}&\textbf{.617}&.605&\textbf{.625}&\textbf{.575}&\textbf{.589}&\textbf{.568}&\textcolor{gray}{.0538} \\

    \bottomrule
    \end{tabular}
    \caption{Story-level Pearson ($\bm{r_p}$) / Spearman ($\bm{r_s}$) correlations and average API cost per sample (\$$\bm{/item}$) of on HANNA benchmark. We implemented and tested all the methods with p-value < 0.05.}
    \label{tab: Main results3}
\end{table*}

\begin{table}[t]
    \renewcommand\arraystretch{1.3}
    \small
    \centering
    \setlength{\tabcolsep}{0.20em} 
    \begin{tabular}{lccccccccc}
    \toprule
    \multirow{2}{*}{\textbf{Method}} &\multicolumn{2}{c}{\textbf{QAGS-C}} & & \multicolumn{2}{c}{\textbf{QAGS-X}} & & \multicolumn{3}{c}{\ \ \ \ \ \ \textbf{Average}\ \ \ \ \ \ \ \ \ \ } \\
     & \ \ $\bm{r_p}$\ &\ $\bm{r_s}$\ & & \ \ $\bm{r_p}$\ &\ $\bm{r_s}$& & \ \ $\bm{r_p}$\ \ &\ $\bm{r_s}$\ \ \ & \ \$$\bm{/item}$ \\   
    \hline
    \multirow{1}{*}{BERTScore$^*$} &   .576&.505& & .024&.008& & .300&.256&- \\
    \multirow{1}{*}{QAGS$^*$} &   .545&-& & .175&-& & .375&-&- \\
    \multirow{1}{*}{G-Eval$^*$} &   .631&\textbf{.685}& & .558&.537& & .599&.611&- \\
    \multirow{1}{*}{CloserLook} &    .581&.602& & .549&.573& & .498&.478&\textcolor{gray}{.0691} \\
    \multirow{1}{*}{\textsc{BatchEval}} &    \textbf{.785}&.643& & \textbf{.618}&\textbf{.634}& & \textbf{.682}&\textbf{.639}&\textcolor{gray}{.0521} \\

    \bottomrule
    \end{tabular}
    
    \caption{Results on QAGS benchmark (QAGS with -C and -X denote subset CNN and XSUM respectively). The results of methods with $^*$ come from G-EVAL. Our replication results of G-Eval are lower than those reported in the original paper, so we present the original results here to avoid potential replication errors. }
    \label{tab: Main results4}
    \vspace{-0.5cm}
\end{table}

\subsection{Overall Performance of \textsc{BatchEval}}
\label{overall}
As shown in Table~\ref{tab: Main results},~\ref{tab: Main results2},~\ref{tab: Main results3},~\ref{tab: Main results4}, \textsc{BatchEval} achieves an average of 6.5 points (10.5\%) Pearson and 4.5 points (7.1\%) Spearman correlations improvements with humans across four benchmarks compared to the best performing methods. 
From the perspective of Theorem~\ref{theorem2}, as shown in Figure~\ref{fig:diver}, we found that the reason \textsc{BatchEval} outperforms CloserLook under score ensemble ($Err(\Bar{s},y)$) is twofold. First, \textsc{BatchEval} attains more accurate single predictions ($Err(\mathcal{S},y)$) through thorough in-batch comparison. Second, the scoring diversity ($Var(\mathcal{S})$) of \textsc{BatchEval} is significantly improved. This validates that iterative heterogeneous batch composition strategy can provide LLM with unbiased varying evaluation references, thus stably enhancing diversity and ensemble performance.

\begin{figure*}[!htb]
    \centering
    \subfigure{\includegraphics[width=0.95\hsize]{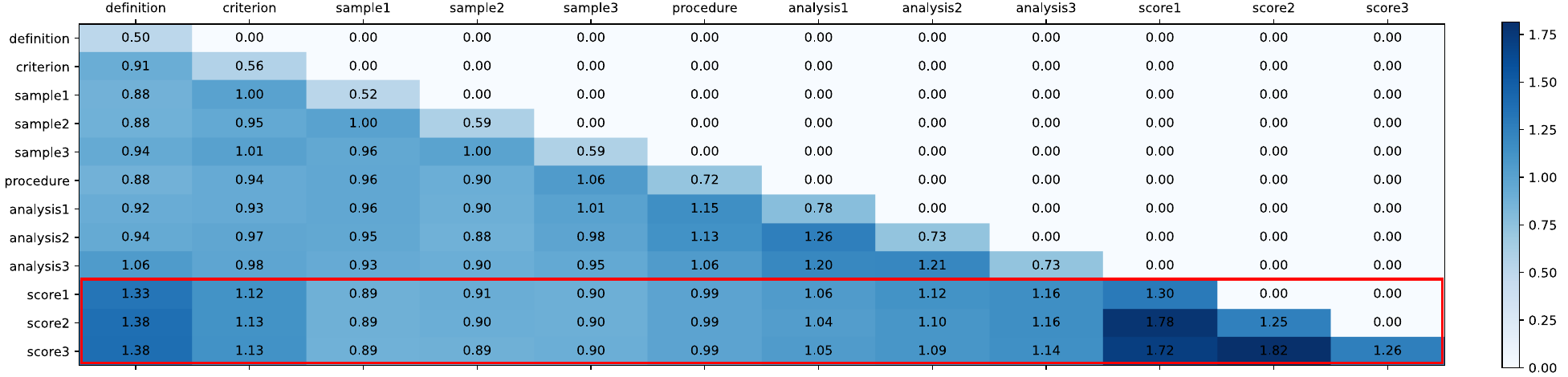}\label{fig: sub_figure1}} \hspace{-0.50cm} \\
    \vspace{-0.30cm}
    \subfigure{\includegraphics[width=0.96\hsize]{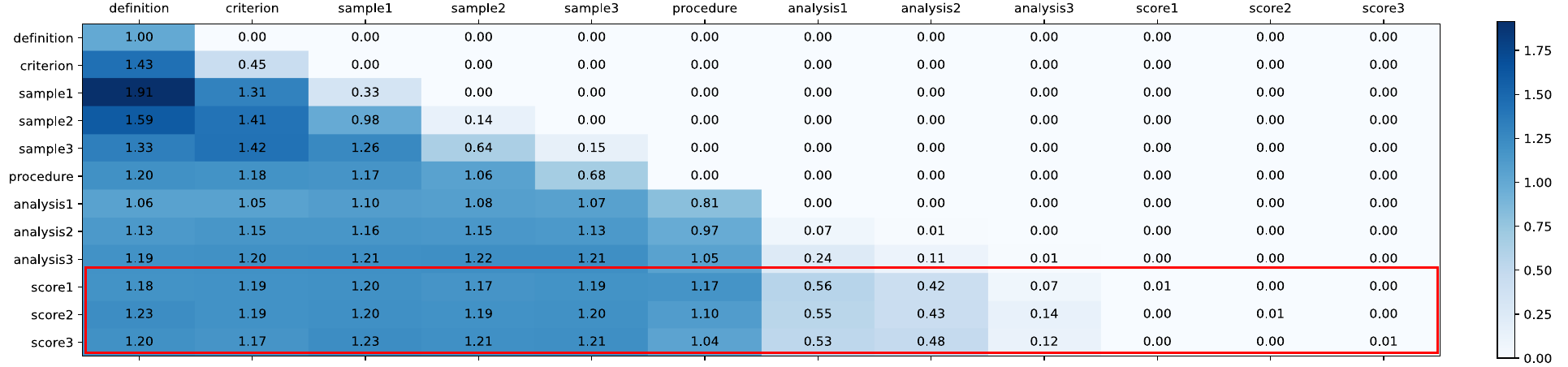}\label{fig: sub_figure2}} \hspace{-0.10cm}
    \setlength{\belowcaptionskip}{0pt}
    \caption{Normalized attention matrices of the first (top figure) and last (bottom figure) transformer layer with Llama-2-70b-chat-hf. We set batchsize as 3 for clear demonstration. See Appendix~\ref{sec:normalize} for the normalizing process.}
    \label{fig:attention}
\end{figure*}

In terms of cost, \textsc{BatchEval} only consumes 64\% API expenses of the best performing baselines. 
This is because we only use the average scores from 5 iterations and allow in-batch samples to share a single prompt, while the LLM-based baselines average scores from 20 generations.\footnote{Due to changes in the prompt during iteration, the prompt expense needs to be billed 5 times for our method, whereas baselines require only once. Therefore the expenditure ratio (64\%) is higher than the proportion of generations (5:20).} Considering that baselines reach ensemble saturation at about 20 generations, \textsc{BatchEval} has broad potential for performance improvement by further increasing the number of iterations.

\subsection{Robustness of \textsc{BatchEval}}
\label{robustness}
\paragraph{Robustness against Prompt Design}
\label{sec:rob_prompt}
We test \textsc{BatchEval} and CloserLook respectively on prompts written by human and rewritten by GPT-4, with results as shown in Table~\ref{tab: Main results2}. We calculate the average difference in correlations across metrics under two types of prompts. The standard deviation of $\bm{r_p}$ and $\bm{r_s}$ are 0.009 and 0.007 for CloserLook, while only 0.006 and 0.002 for \textsc{BatchEval}. This verifies that \textsc{BatchEval} attains better robustness against prompt design by introducing in-batch samples as additional references.

\paragraph{Robustness against Noise}
\label{sec:rob_noise}
As shown in Figure~\ref{fig:dis}, the score distribution of \textsc{BatchEval} is more uniform and has lower entropy compared with CloserLook due to in-batch comparison with decimal scoring format, which can theoretically enhance robustness against noise according to Theorem~\ref{theorem1}. We further experimentally validate this in Appendix~\ref{sec:robustness_noise}.

\subsection{Further Discussion and Analysis}
\label{analysis}
\paragraph{Relationship with In-context-learning} 
ICL \citep{gpt3} can also provide sample-side references by incorporating samples and corresponding answers into the prompt. The main differences between ICL and \textsc{BatchEval} are: (1) \textsc{BatchEval} can provide LLM with varying and comprehensive references through iterative heterogeneous batch, while the references provided by ICL are relatively fixed and may bring bias (sensitive to prompt design). (2) \textsc{BatchEval} uses in-batch samples as references to each other, thus saving the costs of demonstrations in ICL prompts. Thanks to the aforementioned advancements, \textsc{BatchEval} outperforms CloserLook with ICL by more than 5 points Pearson correlations while only incurs 61.8\% expense (Table~\ref{tab: Main results}).

\paragraph{Working Mechanism of \textsc{BatchEval}}
To further understand how \textsc{BatchEval} benefits from in-batch comparison, we visualized the normalized attention matrices of the first and last layers of Llama-2-70b-chat-hf (Figure~\ref{fig:attention}). The value at (X,Y) represents the average normalized attention of tokens corresponding to X towards tokens corresponding to Y. We observe that in the final scoring phase (red box), LLM first perceives samples with varied qualities based on the already generated scores and analyses at the shallower layers. Afterwards, LLM completes scoring based on criterion and comparison between samples at the deeper layers. This process demonstrates the in-batch comparison mechanism of \textsc{BatchEval}, which we hope can inspire future research.

\section{Conclusions}
In this paper, we propose \textsc{BatchEval}, a new text evaluation paradigm that evaluate samples batch-wise to alleviate the limitations of sample-wise evaluation paradigm. We explore variants of \textsc{BatchEval} on multiple dimensions and figure out the optimal settings. Following the human evaluation method, \textsc{BatchEval} treats in-batch samples and criterion as complementary references and optimizes the batch composition through iteration to eliminate batch bias. Comprehensive experiments have confirmed that \textsc{BatchEval} can achieve higher consistency with humans at a lower cost, while also demonstrating better robustness to prompt design and noise. We further analyze and reveal the working mechanism of \textsc{BatchEval}, shedding lights on future work.
\section*{Limitations}
From an objective perspective , we think there are two main limitations of this paper:
\begin{enumerate}

\item \textsc{BatchEval} requires LLMs to have a certain capability to handle longer contexts. From Appendix~\ref{sec:hyper}, we found that as the batchsize increases, LLMs struggle to handle too many samples, leading to a performance decline. We also attempted to test \textsc{BatchEval}'s performance on Llama-2-13b-chat-hf and found that the batchsize must be set to 2 or 3 to see any benefits. Therefore, when setting the batchsize, we cannot exceed the limit of how many samples an LLM can process in a single context. Fortunately, we discovered that a batchsize of 10 is suitable for current mainstream LLMs. Additionally, as LLMs continue to advance, they can handle increasingly larger contexts. Thus, from this perspective, \textsc{BatchEval} is a scalable method that improves alongside the capabilities of LLMs (increasing the batchsize within the capabilities of the LLM can enhance the evaluation effectiveness of the LLM).

\item We only explored a limited number of schemes of \textsc{BatchEval}. We leave exploring possible schemes of \textsc{BatchEval} for future research.

\end{enumerate}

\section*{Ethics Statement}
All of the datasets used in this study were publicly available, and no annotators were employed for our data collection. We confirm that the datasets we used did not contain any harmful content and was consistent with their intended use (research). We have cited the datasets and relevant works used in this study.

\bibliography{acl_latex}

\appendix
\clearpage
\appendix
\label{sec:appendix}

\section{Proof of Theorem \ref{theorem1}}
\label{sec:proofoft1}
For any $f(x)$, the probability density function of score distribution, the Spearman correlation $\mathbb{E}(r_s)$ between the original scores and scores adding a small disturbance has an upper bound:
\begin{equation}
    \mathbb{E}(r_s) \leq 1 - \frac{6\mathbb{E}(\lambda)^2}{n^2-1},
\end{equation}
and the equality condition is $f(x) \equiv 1, \forall x \in [0,1]$.

\newtheorem{proof}{Proof}
\begin{proof}
The ranking difference $d(x)$ before and after disturbance is :
\begin{equation}
d(x) = \int_x^{x+\lambda} f(x) d x
\end{equation}
According to the definition of Spearman correlations, $E(r_s)$ can be written as:
\begin{equation}
\label{formula2}
\mathbb{E}(r_s)=\mathbb{E}(1-\frac{6 \sum_{i=1}^n d(x_i)^2}{n\left(n^2-1\right)}),
\end{equation}
we derive the lower bound of $\mathbb{E}(d(x)^2)$ as follows:

\begin{equation}
\small
\begin{split}
\mathbb{E}(d(x)^2) =& \int_0^1\left(\int_x^{x+\mathbb{E}(\lambda)} f(u) d u\right)^2 f(x) d x \\
=&\int_0^1\left(\int_x^{x+\mathbb{E}(\lambda)} f(u) d u \sqrt{f(x)}\right)^2 d x\\
=&\int_0^1\left(\int_x^{x+\mathbb{E}(\lambda)} f(u) d u \sqrt{f(x)}\right)^2 d x\\ &\cdot \int_0^1 f(x) d x \\
\geq&\left(\int_0^1 \int_x^{x+\mathbb{E}(\lambda)} f(u) d u f(x) d x\right)^2  \\
& (Cauchy's\ Inequality )\\
=&\left(\int_0^1 \mathbb{E}(\lambda) \cdot f(x) \cdot f(x) d x\right)^2(\mathbb{E}(\lambda) \rightarrow 0) \\
=&\mathbb{E}(\lambda)^2\left(\int_0^1 f(x) \cdot f(x) d x\right)^2\\
=&\mathbb{E}(\lambda)^2\left(\int_0^1 f(x)^2 d x \cdot \int_0^1 1^2 d x\right)^2\\
\geq& \mathbb{E}(\lambda)^2\left(\left(\int_0^1 f(x) d x\right)^2\right)^2 \\
& (Cauchy's\ Inequality )\\
=&\mathbb{E}(\lambda)^2
\end{split}
\end{equation}

The equality condition is $f(x) \equiv 1\  for\ x \in [0,1]$. Taking the lower bound of $\mathbb{E}(d(x)^2)$ into Eq.~\eqref{formula2}, 
we conclude the proof.
Note that higher $\mathbb{E}(r_s)$ denotes better robustness against noise. Hence, we can derive that the robustness against noise correlates positively with the uniformity of score distribution.
\end{proof}

\section{Hyperparameter Analysis}
\label{sec:hyper}
In the experiments of the main text, we set the batch size to 10 and the temperature to 0.2. In this section, we explore the impact of different hyperparameter choices on performance.

\begin{figure*}[t]
\centering
\includegraphics[width=0.96\textwidth]{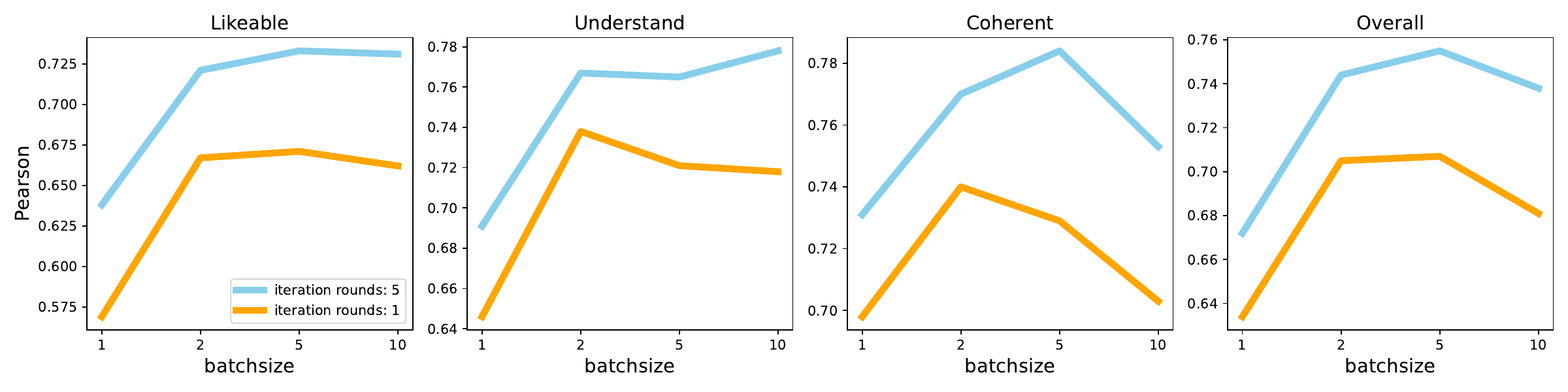} 
\caption{Dialog-level Pearson correlations on FED-dialog dataset of \textsc{BatchEval} with different batchsize.}
\label{fig:batch}
\end{figure*}

\begin{figure*}[t]
\centering
\includegraphics[width=0.96\textwidth]{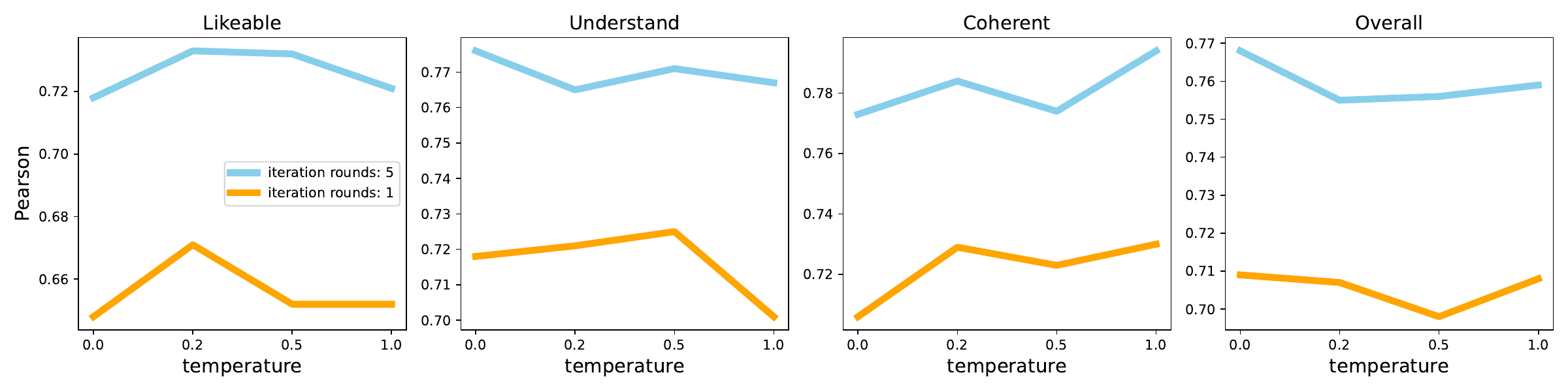} 
\caption{Dialog-level Pearson correlations on FED-dialog dataset of \textsc{BatchEval} with different temperature.}
\label{fig:temp}
\end{figure*}

\subsection{Effect of Batchsize}
On FED dataset, we test \textsc{BatchEval} with batchsize among [1, 2, 5, 10]. As shown in Figure~\ref{fig:batch}, we found that as the batch size increases, the performance generally undergoes a process of initial improvement followed by a decline. Similar observations were made on other datasets as well. We further discovered that the performance turning point of the ensemble results from five iterations is slightly delayed compared to a single prediction. Considering that increasing the batchsize will make the combination of in-batch samples more diverse, thereby increasing scoring diversity, we have the following conjecture about Figure~\ref{fig:batch}: When the batchsize starts to increase from 1, due to the effect of in-batch comparison and the increase in diversity, the performance of both 1-round score and ensemble score increase a lot. However, as the batchsize continues to increase, LLM finds it difficult to handle too many samples simultaneously, resulting in a decrease in 1-round score performance. When the rate of decrease in 1-round score performance gets greater than the rate of increase in diversity, ensemble score performance also begins to decrease according to Theorem~\ref{theorem2}. Therefore, the batchsize should not be too large or too small. We found that setting the batchsize to 10 can achieve superior performance on different tasks. We also believe that for LLMs with weaker ability to handle longer context, the batchsize should be set to be smaller. Fortunately, we have noticed that current LLMs are continually improving in processing long contextual texts, which illuminates further development prospects for \textsc{BatchEval} in the future.

\subsection{Effect of Temperature}
We also test \textsc{BatchEval} with temperature among [0, 0.2, 0.5, 1]. We found that as the temperature rises in Figure~\ref{fig:temp}, the performance of \textsc{BatchEval} does not exhibit a uniform trend of change. Overall, the performance of 5 iterations is relatively stable along the temperature dimension, suggesting that \textsc{BatchEval} is quite robust to temperature variations.

\begin{table*}[t]
    \renewcommand\arraystretch{1.3}
    \small
    \centering
    \setlength{\tabcolsep}{0.15em} 
    \begin{tabular}{lccccccccccc}
    \toprule
    \multirow{2}{*}{\textbf{Method}} &\multicolumn{2}{c}{\ \textbf{Likeable}\ \ \ \ } & \multicolumn{2}{c}{\ \textbf{Understand}\ \ \ \ } & \multicolumn{2}{c}{\ \ \textbf{Coherent}\ \ \ \ \ \ }& \multicolumn{2}{c}{\ \ \textbf{Overall}\ \ \ \ \ \ }&  \multicolumn{3}{c}{\ \ \ \ \ \textbf{Average}\ \ \ \ \ \ \ \ \ } \\
     & \ \ $\bm{r_p}$&\ $\bm{r_s}$&\ \ $\bm{r_p}$&\ $\bm{r_s}$&\ \ $\bm{r_p}$&$\bm{r_s}$\ \ &\ \ $\bm{r_p}$&$\bm{r_s}$\ \ &\ \ $\bm{r_p}$\ \ \ \ \ \ \ \ \ &$\bm{r_s}$\ \ \ \ \ \ \ \ & \$$\bm{/item}$ \\   
    \hline

    \multirow{1}{*}{CloserLook \textit{w/o} noise} & .658&.680&.701&.614&.739&.755&.715&.684&.703\ \ \ \ \ \ \ \ \ &.683\ \ \ \ \ \ \ \ \ &\textcolor{gray}{.0785} \\
    \multirow{1}{*}{CloserLook \textit{w} noise} & .509 &.580&.626&.606&.608&.605&.632&.616&.594\ \tiny\textcolor{red}{(-.109)}&.602\ \tiny\textcolor{red}{(-.081)}&\textcolor{gray}{.0866} \\
    \cdashline{1-12}
    \multirow{1}{*}{\textsc{BatchEval} \textit{w/o} noise} &.731&.741&.778&.696&.753&.757&.738&.729&.750\ \ \ \ \ \ \ \ \ &.731\ \ \ \ \ \ \ \ \ &\textcolor{gray}{.0314} \\
    \multirow{1}{*}{\textsc{BatchEval} \textit{w} noise} &.729&718&.775&.700&.764&.754&.720&.724&.747\ \tiny\textcolor{red}{(-.003)}&.724\ \tiny\textcolor{red}{(-.007)}&\textcolor{gray}{.0344}\\

    \bottomrule
    \end{tabular}
    \caption{Story-level Pearson ($\bm{r_p}$) / Spearman ($\bm{r_s}$) correlations and average API cost per sample (\$$\bm{/item}$) of on HANNA benchmark. We tested all the methods for a fair comparison with p-value < 0.05.}
    \label{tab:noise}
\end{table*}

\section{Robustness against Noise}
\label{sec:robustness_noise}
To test the robustness against noise of \textsc{BatchEval}, we use an external tool\footnote{nlpaug(\url{https://github.com/makcedward/nlpaug})} to add noise to the input and calculate the changes in performance before and after the noise is added. For the sake of noise balance, we randomly replace 5\% of tokens with synonyms and randomly delete 5\% of tokens. As shown in Table~\ref{tab:noise}, CloserLook experiences a decrease of 0.109 in Pearson correlation and 0.081 in Spearman correlation, respectively. In contrast, \textsc{BatchEval} only shows a decrease of 0.003 and 0.009, respectively. This indicates that \textsc{BatchEval} has much better robustness to noise.

\begin{table*}[ht]
    \renewcommand\arraystretch{1.3}
    \small
    \centering
    \setlength{\tabcolsep}{0.10em} 
    \begin{tabular}{lcrcrcrcrcrcrrc}
    \toprule
    \multirow{2}{*}{\textbf{Method}} & \multirow{2}{*}{\textbf{Scheme}\ } & \multicolumn{2}{c}{\textbf{\ \ Engaging\ \ }} & \multicolumn{2}{c}{\textbf{Understand}} & \multicolumn{2}{c}{\ \textbf{Naturalness}\ } & \multicolumn{2}{c}{\ \ \textbf{Coherence}\ \ }& \multicolumn{2}{c}{\ \ \ \textbf{Overall}\ \ \ \ }& \multicolumn{3}{c}{\ \ \ \ \ \ \ \ \ \ \textbf{Average}\ \ \ \ \ \ \ \ \ \ } \\
     & & $\bm{r_p}$\ &\ $\bm{r_s}$&$\bm{r_p}$\ &\ $\bm{r_s}$&$\bm{r_p}$\ &\ $\bm{r_s}$&$\bm{r_p}$\ &\ $\bm{r_s}$&$\bm{r_p}$\ &\ $\bm{r_s}$&$\bm{r_p}$\ &$\bm{r_s}$\ \ & \$$\bm{/item}$ \\   
    \hline

    \multirow{3}{*}{\textsc{BatchEval}} &default & \textbf{.792}&\textbf{.790} & .694&.727 & \textbf{.730}&.\textbf{735} & \textbf{.740}&.744  & .805&.800 & \textbf{.752}&.759 & \textcolor{gray}{.0529}\\
   & 3 stage & .782&.778&.667&.725&.712&704&.712&.714&.797&.798&.734&.744&\textcolor{gray}{.0541}\\
   & 3 stage w/o rank results &.789&.785&\textbf{.701}&\textbf{.733}&.721&.727&.735&\textbf{.747}&\textbf{.810}&\textbf{.808}&.751&\textbf{.760}&\textcolor{gray}{-}
\\
    \bottomrule
    \end{tabular}
    \caption{Comparison of \textsc{BatchEval} with different scheme. \textit{3 stage w/o ranking results} means results of deleting the ranking and scoring contents of LLM's three stage procedure response and asking LLM to score based on the remaining contents (samples and analyses)}
    \label{tab: Main results5}
\end{table*}

\section{Inferior Performance of Three Stage Procedure}
\label{sec:rank}
As shown in Table~\ref{tab: Main results}, we observe a performance drop of \textsc{BatchEval} with three stage procedure, though it may be closer to human evaluation procedure. We speculate this may be due to the LLM’s over-reliance on ranking results while neglecting the analyses and samples during scoring. To validate this, we delete the ranking and scoring contents of LLM's three stage procedure response and ask LLM to score based on the remaining contents (samples and analyses). If the new scoring results perform similarly to \textsc{BatchEval} with two stage procedure, the inferior performance of \textsc{BatchEval} with three stage procedure can be attributed to its excessive focus on ranking results. Otherwise, the reason lies in the decrease in the quality of analyses. As shown in Table~\ref{tab: Main results5}, the performance of three stage w/o rank results is on par with that of two stage procedure. This validates our conjecture that the over-reliance on ranking results causes the performance drop of \textsc{BatchEval} with three stage procedure.

\section{Relationship with Pair-wise Evaluation}
The current mainstream text evaluation approach adopts sample-wise assessment. Alternatively, an LLM evaluator is presented with a question and two answers, and is tasked with determining which one is better or declaring a tie \citep{judging,alpacafarm}. However, as the number of models to be evaluated grows, the scalability of pairwise comparison becomes a challenge, due to the quadratic increase in the potential number of pairs. Therefore, this pair-wise paradigm has not been as extensively studied as sample-wise evaluation. \citet{judging} validates that this method performs slightly better than a sample-wise evaluator, potentially due to its ability to discern subtle differences between specific pairs. 

Similarly, we have enhanced the evaluation capabilities of the LLM evaluator through in-batch sample comparison. The main difference lies in the composition of our batches, which consist of different samples rather than responses from different models to the same sample, thereby offering good scalability.

\section{Batch Composition Strategies}
\label{sec:batchstrategydetail}
\subsection{Homogenized Batch}
Given scores $s^{1:|\mathcal{D}|}$ for samples $x^{1:|\mathcal{D}|}$ predicted by LLM in the previous round, we first sort the scores and attain the corresponding indexes $index^{1:|\mathcal{D}|}$. Based on this, we get indexes of homogenized batch $b^i = index^{1+(i-1)*10:i*10}$.

\subsection{Heterogenized Batch}
Given scores $s^{1:|\mathcal{D}|}$ for samples $x^{1:|\mathcal{D}|}$ predicted by LLM in the previous round, we first sort the scores and attain the corresponding indexes $index^{1:|\mathcal{D}|}$. Considering that our default batchsize is 10, we group the indexes into 10 splits $split^{1:10}$, where $split^i = index^{1+(i-1)\times \lceil \frac{|\mathcal{D}|}{10} \rceil:i\times \lceil \frac{|\mathcal{D}|}{10} \rceil}$. Based on this, we get indexes of heterogenized batch $b^i = \{split^{j,i}|j \in [1,10]\}$.

\section{Introduction of Baselines}
\label{sec:baselines}
\subsection{Rule-based Methods}
\paragraph{BLEU} \citep{bleu} BLEU is a renowned metric for measuring word overlap, which evaluates n-gram precision in a generated sequence against a reference. It includes a brevity penalty to counteract its inherent preference for shorter sentences, ensuring a more comprehensive assessment.
\paragraph{METEOR} \citep{meteor} is an advancement over BLEU, utilizing a harmonic mean of precision and recall, and also incorporating stemming and synonym use in its evaluation.

\subsection{Embedding-based Methods}
\paragraph{Vector Extrema} \citep{vextrema} is a scoring method that uses cosine similarity between sentence embeddings, identifying the highest value in each dimension of the word embedding for evaluation.
\paragraph{BERTScore} \citep{bertscore} is a method that utilizes a pretrained BERT \citep{bert} model to optimally align each word in a reference response with a single word in the generated sequence. By doing so, BERTScore computes the recall of the generated sequence.

\subsection{Learning-based Methods}
\paragraph{USR} \citep{usr} is a dialogue response evaluation method that uses one masked language model and two dialogue retrieval models to assess various sub-qualities of a sample and then integrates these evaluations into a comprehensive overall score.

\paragraph{BCR} \citep{bcr} is a dialogue response evaluation method that use a dynamic loss function to train a BERT model with uniform score distribution.

\paragraph{FED} \citep{fed} is a unified dialogue evaluation method that uses pretrained language models to calculate scores based on the difference in the probability of generating positive and negative evaluation words for a certain criterion.

\paragraph{DynaEval} \citep{dynaeval} is also a unified dialogue evaluation method that leverages graph convolutional network to model the sentences among a dialogue for accurate evaluation.

\paragraph{QAGS} \citep{qags} is a method that based on question-answering, which creates questions from a summary and then verifies whether their answers are present in the original source document.

\section{Details of Normalizing Process}
\label{sec:normalize}
We will introduce how to normalize the attention matrix to make it more visually appealing like in Figure~\ref{fig:attention}.
Due to the autoregressive generation mode of mainstream LLMs, the expected values of attention between token pairs at different positions vary. If we use $Att(x,y)$ to represent the attention of the $x^{th}$ token to the $y^{th}$ token, then its expected value is $\frac{1}{x}$. Since tokens at different positions will be visualized into the same graph, we first multiply each $Att(x,y)$ by $x$ to make its expected value 1. On this basis, we determine the token intervals corresponding to different strings through word matching, and calculate 
$Att(string1,string2)$ as follows:
\begin{equation}
Att(s1,s2)=Avg(\{Att(x,y)|x\in{s1},y\in{s2}\})
\end{equation}
according to which we plot our attention matrices.

\section{Example Prompts}
\label{sec:prompts}
\subsection{Evaluate Coherence for Topical-Chat}

\paragraph{default prompt}
\begin{quote}
{\itshape
You will be given a batch of \{\{number\}\} samples. Each sample contains a conversation between Speaker A and Speaker B and one potential response for the next turn.

Your task is to assign a float score to the response on one metric.

You should carefully horizontally compare the given samples in order to assign a suitable float score to each sample.

Please make sure you read and understand these instructions carefully. Please keep this document open while reviewing, and refer to it as needed.
\\
\\
Evaluation Criteria:
\\
\\
Coherence (floating point numbers within the interval [1,3]): Does the response serve as a valid continuation of the conversation history?

- A float score near 1 (no) means that the response drastically changes topic or ignores the conversation history.

- A float score near 2 (somewhat) means the response refers to the conversation history in a limited capacity (e.g., in a generic way) and shifts the conversation topic.

- A float score near 3 (yes) means the response is on topic and strongly acknowledges the conversation history.
\\
\\
Conversations and corresponding potential response to be evaluated:
\\
\\
\{\{Data\}\}

Evaluation Form (Answer by starting with "I will do my best to provide individual analysis for each sample. Analysis:" to analyze the given samples regarding the evaluation criteria as concise as possible (Attention: Don't give your scores during this step). After analysing all the samples, please give all the float scores in order following the template "Float Scores: [Sample1:score of Sample1,...,Sample\{\{number\}\}:score of Sample\{\{number\}\}]".

- Coherence:

}
\end{quote}

\paragraph{one stage prompt}
\begin{quote}
{\itshape
You will be given a batch of \{\{number\}\} samples. Each sample contains a conversation between Speaker A and Speaker B and one potential response for the next turn.

Your task is to assign a float score to the response on one metric.

You should carefully horizontally compare the given samples in order to assign a suitable float score to the given samples one by one.

Please make sure you read and understand these instructions carefully. Please keep this document open while reviewing, and refer to it as needed.
\\
\\
Evaluation Criteria:

Coherence (floating point numbers within the interval [1,3]): Does the response serve as a valid continuation of the conversation history?

- A float score near 1 (no) means that the response drastically changes topic or ignores the conversation history.

- A float score near 2 (somewhat) means the response refers to the conversation history in a limited capacity (e.g., in a generic way) and shifts the conversation topic.

- A float score near 3 (yes) means the response is on topic and strongly acknowledges the conversation history.
\\
\\
Conversations and corresponding potential response to be evaluated:

\{\{Data\}\}
\\
\\
Evaluation Form (Answer by starting with "I will do my best to provide individual analysis and give a suitable float score for each sample in order". When rating for each sample, please follow the template "Score of SampleX:[float score]").

- Coherence:

}
\end{quote}

\paragraph{three stage prompt}
\begin{quote}
{\itshape

You will be given a batch of \{\{number\}\} samples. Each sample contains a conversation between Speaker A and Speaker B and one potential response for the next turn.

You will be introduced to a metric to be evaluated.

You should carefully horizontally compare the given samples in order to assign a suitable float score to each sample.

Please make sure you read and understand these instructions carefully. Please keep this document open while reviewing, and refer to it as needed.
\\
\\
Evaluation Criteria:

Coherence (floating point numbers within the interval [1,3]): Does the response serve as a valid continuation of the conversation history?

- A float score near 1 (no) means that the response drastically changes topic or ignores the conversation history.

- A float score near 2 (somewhat) means the response refers to the conversation history in a limited capacity (e.g., in a generic way) and shifts the conversation topic.

- A float score near 3 (yes) means the response is on topic and strongly acknowledges the conversation history.
\\
\\
Conversations and corresponding potential response to be evaluated:

\{\{Data\}\}
\\
\\
Answer by starting with "I will do my best to provide individual analysis for each sample. Analysis:" to analyze the given samples regarding the evaluation criteria as concise as possible (Attention: Don't give your scores during this step).
After analysing all the samples, please horizontally compare the given samples, rank all the samples according to the analysis of the response and give the corresponding reasons.
After ranking, according to the analysis and rank, please give all the float scores in order following the template "Float Scores: [Sample1:score of Sample1,...,Sample\{\{number\}\}:score of Sample\{\{number\}\}]".

- Coherence:
}
\end{quote}

\paragraph{Integer prompt}
\begin{quote}
{\itshape

You will be given a batch of \{\{number\}\} samples. Each sample contains a conversation between Speaker A and Speaker B and one potential response for the next turn.

Your task is to rate the responses on one metric.

You should carefully horizontally compare the given samples in order to assign a score to each sample.

Please make sure you read and understand these instructions carefully. Please keep this document open while reviewing, and refer to it as needed.
\\
\\
Evaluation Crieteria:

Coherence (1-3): Does the response serve as a valid continuation of the conversation history?

- A score of 1 (no) means that the response drastically changes topic or ignores the conversation history.

- A score of 2 (somewhat) means the response refers to the conversation history in a limited capacity (e.g., in a generic way) and shifts the conversation topic.

- A score of 3 (yes) means the response is on topic and strongly acknowledges the conversation history.
\\
\\
Conversations and corresponding potential response to be evaluated:

\{\{Data\}\}
\\
\\
Evaluation Form (Answer by starting with "I will do my best to provide individual analysis for each sample. Analysis:" to analyze the given samples regarding the evaluation criteria as concise as possible (Attention: Don't give your scores during this step). After analysing all the samples, please give all the scores in order following the template "Scores: [Sample1:score of Sample1,...,Sample\{\{number\}\}:score of Sample\{\{number\}\}]".

- Coherence:
}
\end{quote}

\subsection{Evaluate Coherent for FED-Dialogue}

\paragraph{default prompt}
\begin{quote}
{\itshape

You will be given a batch of \{\{number\}\} samples. Each sample contains a conversation between User and a dialogue System.

Your task is to assign a float score to the sample on one metric.

You should carefully horizontally compare the given samples in order to assign a suitable float score to each sample.

Please make sure you read and understand these instructions carefully. Please keep this document open while reviewing, and refer to it as needed.
\\
\\
Evaluation Criteria:

Coherent (floating point numbers within the interval [1,3]): Does System maintain coherence and a good flow of conversation throughout the dialogue?

- A float score near 1 (not coherent) means that System's responses are unrelated to the conversation topic and may disrupt or confuse the flow of the dialogue.

- A float score near 2 (somewhat coherent) means that System's responses are partially related to the conversation topic but may not be clear or direct.

- A float score near 3 (very coherent) means that System's responses are closely related to the conversation topic and contribute to maintaining a smooth dialogue.
\\
\\
Conversations to be evaluated:

\{\{Data\}\}
\\
\\
Evaluation Form (Answer by starting with "I will do my best to provide individual analysis for each sample. Analysis:" to analyze the given samples regarding the evaluation criteria as concise as possible (Attention: Don't give your scores during this step). After analysing all the samples, please give all the float scores in order following the template "Float Scores: [Sample1:score of Sample1,...,Sample\{\{number\}\}:score of Sample\{\{number\}\}]".

- Coherent:

}
\end{quote}

\subsection{Evaluate Coherence for HANNA}

\paragraph{default prompt}
\begin{quote}
{\itshape

You will be given a batch of \{\{number\}\} samples. Each sample contains a prompt and a story generated following the prompt.

Your task is to assign a float score to the story according to the prompt on one metric.

You should carefully horizontally compare the given samples in order to assign a suitable float score to each sample.

Please make sure you read and understand these instructions carefully. Please keep this document open while reviewing, and refer to it as needed.
\\
\\
Evaluation Criteria:

Coherence (floating point numbers within the interval [1,5]) Measures whether the story makes sense?

- A float score near 1 means the story does not make sense at all. For instance, the setting and/or characters keep changing, and/or there is no understandable plot.

- A float score near 2 means most of the story does not make sense.

- A float score near 3 means the story mostly makes sense but has some incoherences.

- A float score near 4 means the story almost makes sense overall, except for one or two small incoherences.

- A float score near 5 means the story makes sense from beginning to end.
\\
\\
Prompts and corresponding stories to be evaluated:

\{\{Data\}\}
\\
\\
Evaluation Form (Answer by starting with "I will do my best to provide individual analysis for each sample. Analysis:" to analyze the given samples regarding the evaluation criteria as concise as possible (Attention: Don't give your scores during this step). After analysing all the samples, please give all the float scores in order following the template "Float Scores: [Sample1:score of Sample1,...,Sample\{\{number\}\}:score of Sample\{\{number\}\}]".

- Coherence:

}
\end{quote}

\subsection{Evaluate Factual Consistency for QAGS}

\paragraph{default prompt}
\begin{quote}
{\itshape

You will be given a batch of \{\{number\}\} samples. Each sample contains an article and a sentence.

Your task is to determine if the sentence is factually correct given the contents of the article.

You should carefully horizontally compare the given samples in order to assign a suitable float score to each sample.

Please make sure you read and understand these instructions carefully. Please keep this document open while reviewing, and refer to it as needed.
\\
\\
Evaluation Criteria:

Consistency ([1,3]) - Is the sentence supported by the article? (consistent with the article)

- A float score near 1 (not) means that the sentence is totally not supported by the article.

- A float score near 2 (somewhat) means that the sentence is partially supported by the article.

- A float score near 3 (very) means that the sentence is completely supported by the article.
\\
\\
Articles and corresponding sentences to be evaluated:

\{\{Data\}\}
\\
\\
Evaluation Form (Answer by starting with "I will do my best to provide individual analysis for each sample. Analysis:" to analyze the given samples regarding the evaluation criteria as concise as possible (Attention: Don't give your scores during this step). After analysing all the samples, please give all the float scores in order following the template "Float Scores: [Sample1:score of Sample1,...,Sample\{\{number\}\}:score of Sample\{\{number\}\}]".

- Consistency:

}
\end{quote}

\end{document}